\documentclass[11pt]{article}

\usepackage[preprint]{acl}

\usepackage{times}
\usepackage{latexsym}

\usepackage[T1]{fontenc}
\usepackage[utf8]{inputenc}

\usepackage{microtype}

\usepackage{inconsolata}

\usepackage{graphicx}

\usepackage{xcolor}
\usepackage{placeins}  % in preamble
\usepackage{makecell}
\usepackage{tabularx}   
\usepackage{courier}
\usepackage{amsmath}
\usepackage{subcaption}
\usepackage{multirow}
\usepackage{booktabs}
\usepackage{amssymb} 
\usepackage{bm}
\usepackage{enumitem}   % for customizing itemize spacing (optional, but useful)

\usepackage{helvet}  % DO NOT CHANGE THIS
\usepackage{courier}  % DO NOT CHANGE THIS
\usepackage{url}
\usepackage{natbib}  % DO NOT CHANGE THIS AND DO NOT ADD ANY OPTIONS TO IT
\usepackage{caption} % DO NOT CHANGE THIS AND DO NOT ADD ANY OPTIONS TO IT
\usepackage{algorithm}
\usepackage{algorithmic}
\usepackage{xcolor}

\newcommand{\stitle}[1]{\noindent\textup{\textbf{#1}}}
\newcommand{\ours}{APP}

\title{Evaluating Sparse Autoencoders for Monosemantic Representation}

\author{
  \textbf{Moghis Fereidouni},
  \textbf{Muhammad Umair Haider},
  \textbf{Peizhong Ju},
  \textbf{A.B. Siddique},
\\
  University of Kentucky,
\\
  \small{
    moghis.fereidouni@uky.edu, muhammadumairhaider@uky.edu, 
    peizhong.ju@uky.edu,
    siddique@cs.uky.edu
  }
}

\begin{document}
\maketitle

\begin{abstract}
%\umaircomment{highlighting our controlling approach alot might confuse the reviewer. more focus should be on analysis itself}

A key barrier to interpreting large language models is polysemanticity, where neurons activate for multiple unrelated concepts.
Sparse autoencoders (SAEs) have been proposed to mitigate this issue by transforming dense activations into sparse, more interpretable features. 
While prior work suggests that SAEs promote monosemanticity, no quantitative comparison has examined how concept activation distributions differ between SAEs and their base models. 
%While prior work suggests that SAEs promote monosemanticity, there has been no quantitative comparison with their base models regarding how concept activations are distributed across neurons.
This paper provides the first systematic evaluation of SAEs against base models through activation distribution lens.
%concerning monosemanticity in terms of the separability of concept activation distributions across neurons.
We introduce a fine-grained concept separability score based on the Jensen–Shannon distance, which captures how distinctly a neuron's activation distributions vary across concepts. Using two large language models (Gemma-2-2B and DeepSeek-R1) and multiple SAE variants across five datasets (including word-level and sentence-level), we show that SAEs reduce polysemanticity and achieve higher concept separability. 
To assess practical utility, we evaluate concept-level interventions using two strategies: full neuron masking and partial suppression. We find that, compared to base models, SAEs enable more precise concept-level control when using partial suppression.
Building on this, we propose Attenuation via Posterior Probabilities (APP), a new intervention method that uses concept-conditioned activation distributions for targeted suppression. APP achieves the smallest perplexity increase while remaining highly effective at concept removal \footnote{Code will be released publicly after paper acceptance.}.

\end{abstract}

%\vspace{-20pt}
\section{Introduction}

%\fix{Please read the Introduction Section carefully. I have tried to say everything, we wanted to say. Check everything for correctness. Also, citations...}

Large language models (LLMs) have achieved remarkable performance across a wide range of natural language tasks, often matching or surpassing human-level performance~\cite{luo2025large,achiam2023gpt,touvron2023llama,guo2025deepseek,eslamian2025tabnsa}. 
Nonetheless, understanding how these models internally represent and manipulate concepts remains a major challenge. 
A key obstacle is polysemanticity; the phenomenon where individual neurons respond to multiple, semantically distinct concepts rather than encoding single, interpretable features~\cite{janiak2023polysemantic,olah2017feature,nguyen2016multifaceted}. 
This entanglement complicates the interpretation and analysis of model behavior, posing a significant barrier to building transparent and controllable AI systems~\cite{sharkey2025open,marshall2024understanding,bereska2024mechanistic}.

Dictionary learning via sparse autoencoders (SAEs)~\cite{huben2024sparse,gao2025scaling} has recently emerged as a promising approach to mitigating polysemanticity in neural representations. 
SAEs aim to transform dense activations of a desired component of the base LLM into sparse features by enforcing sparsity and encouraging each neuron to specialize in distinct, concept-specific features~\cite{huben2024sparse,rajamanoharan2024improving,rajamanoharan2024jumping,gao2025scaling}. 
The goal is to produce monosemantic representations, where individual neurons respond to single, well-defined concepts~\cite{huben2024sparse,rajamanoharan2024improving,rajamanoharan2024jumping,gao2025scaling}. 
The underlying hypothesis is intuitive; if we can force the model to use fewer neurons simultaneously, each active neuron should correspond to a more distinct and interpretable concept. 
Empirical studies have shown that SAEs can uncover interpretable features across domains such as vision and language, facilitating improved interpretability~\cite{Shu_Survey_SAE,huben2024sparse,pach2025sparse}.

Most existing evaluations of SAE interpretability are qualitative, relying on case studies or anecdotal neuron visualizations that provide limited systematic insight~\cite{kissane2024interpreting,li2025interpretability}.
The remaining quantitative efforts mainly examine whether neurons are active for given concepts, rather than capturing the full distributional structure of concept activations across neurons~\cite{minegishi2025rethinking, karvonen2025saebench}.

%While SAEs have higher computational costs and degraded downstream performance compared to their dense counterparts~\cite{huben2024sparse}, no work has quantitatively compared their internal representations for monosemanticity: a core requirement for enhancing model interpretability. 
%Without such comparison, the practical and theoretical value of sparsity remains poorly understood.

In this work, we conduct a systematic investigation into the effectiveness of SAEs in promoting monosemanticity in the internal representations of LLMs through the \emph{distributional lens}.
%In contrast to prior work that evaluates SAEs in isolation, we compare their representations to those of the corresponding base models, providing the quantitative assessment of how sparsity affects concept disentanglement.
Particularly, we conduct comprehensive evaluations using two large language models (Gemma-2-2B~\cite{morganeGemma} and DeepSeek-R1~\cite{guo2025deepseek}) and various SAEs of different widths and sparsity levels on five benchmark datasets, including both word-level (POS tagging and NER) and sentence-level (e.g., AG News) tasks.
We begin by quantifying polysemanticity using overlap statistics, measuring the fraction of salient neurons that respond to multiple, semantically distinct concepts. 
While SAEs exhibit lower polysemanticity than their base models, this overlap-based analysis remains coarse-grained; it treats all neurons that respond to multiple concepts as equally entangled, without considering how their activations vary across those concepts. 
In practice, a neuron may activate for several concepts, yet do so with clearly distinct activation distributions, suggesting behavior that may still be considered monosemantic. 
That is, monosemanticity is not solely about binary activation overlap, but rather about the separability of a neuron's activation distributions across concepts.

We formalize this view by introducing a new concept separability score, based on the Jensen–Shannon distance~\cite{lin1991divergence}. 
This fine-grained, distribution-aware metric quantifies how well a neuron's activations separate across different concepts by measuring the distance between their activation distributions.
Using this score, we find that SAEs exhibit higher concept separability than their dense counterparts.

To further evaluate the practical utility of monosemantic representations of SAEs, we examine their effectiveness in enabling concept-level model interventions. 
Specifically, we assess how precisely concept-related behavior can be suppressed in SAEs compared to base models. 
We evaluate two intervention strategies; full neuron masking, which suppresses all activations of salient neurons associated with a target concept, and partial suppression, which intervenes in activations selectively based on their distributional association with the concept. Across SAEs and the base model, partial suppression outperforms full masking in most cases, achieving more effective suppression of the target concept while better preserving unrelated model behavior. Furthermore, our results show that SAEs support more precise and effective concept removal than their dense counterparts, especially when applying partial suppression methods. These findings reinforce the idea that concept separability, when defined in terms of activation distributions rather than binary activation overlap, offers better model control and interpretability.

Additionally, motivated by the varying separability of concept activations across neurons, we introduce Attenuation via Posterior Probabilities (APP), a new intervention method that leverages concept-conditioned activation distributions to selectively suppress target concepts with minimal side effects. Specifically, APP computes the posterior probability that a given activation corresponds to a target concept and attenuates it accordingly.
Among all methods evaluated, APP achieves the smallest degradation in language modeling quality (lowest perplexity increase) while remaining highly competitive with other baselines in targeted concept removal across both SAEs and base models.

%Finally, we show that JS-based separability strongly predicts erasability; neurons with higher JS distance across concepts enable more precise and less disruptive concept erasure.

In summary, this work makes the following contributions:
\vspace{-5pt}
\begin{itemize}
  \item We present the first quantitative analysis of monosemantic representations in SAEs relative to their base LLM through distributional lens.
  \vspace{-5pt}
  \item We introduce a concept separability score, based on the Jensen-Shannon distance, a fine-grained, distribution-aware metric that captures how well neuron activations separate for different concepts.
  \vspace{-5pt}
  \item We propose a new intervention method, which is the least invasive concept erasure technique and highly competitive with existing methods in removing the targeted concept. % selective and
\end{itemize}

\section{Preliminaries}

\stitle{Neuron.} A neuron refers to a component of a hidden state vector in a transformer layer. Given a hidden state $h^l \in \mathbb{R}^d$ at layer $l$, the $j$-th neuron is denoted by $h^l_j$.

\stitle{Concept.} A concept $c_i \in C$ is a semantic category assigned to each input (or its components), where $C = {c_1,\dots,c_k}$. For instance, sentence-level types (e.g., declarative, interrogative) or word-level tags (e.g., noun, verb) can serve as concepts. In this work, we focus both on \emph{sentence-level} and \emph{word-level} concepts.

\stitle{Datasets and Models.} We use five datasets: Part-of-Speech Tagging (POS)~\cite{Pasini_Raganato_Navigli_2021}, Named Entity Recognition (NER)~\cite{ding-etal-2021-nerd}, AG News~\cite{NIPS2015_250cf8b5}, Emotions~\cite{saravia-etal-2018-carer}, and DBpedia~\cite{NIPS2015_250cf8b5}. Our analysis is conducted using two large language models: (1) the Gemma-2-2B model~\cite{morganeGemma} along with its corresponding Sparse Autoencoders (SAEs) from GemmaScope~\cite{lieberum-etal-2024-gemma}, and (2) the DeepSeek-R1 model~\cite{guo2025deepseek} with its associated SAE from LlamaScope~\cite{he2024llama}.

\begin{figure}[!t]
  \centering
  \includegraphics[width=\linewidth]{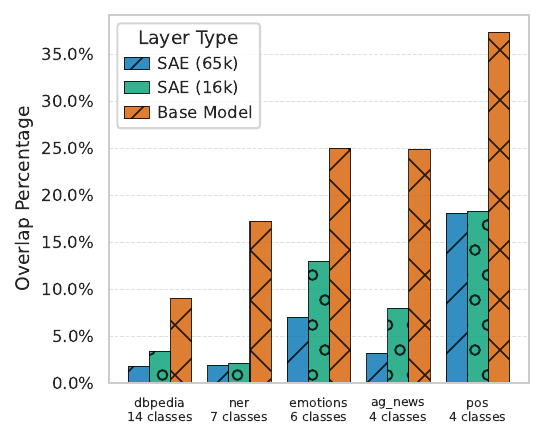}
  \caption{SAEs reduce neuron overlap in comparison to the base model, indicating lower polysemanticity. 
  Higher-capacity SAEs (65k) further reduce overlap, suggesting more effective assignment of distinct neurons to separate concepts.}
  \vspace{-20pt}
  \label{fig:overlap_percentage}
\end{figure}

\vspace{-6pt}
\section{Analyzing Polysemanticity in SAEs}

A neuron is considered polysemantic when it responds to multiple, distinct concepts rather than a single, well-defined one. Several studies have demonstrated that such polysemantic behavior is common in neural networks \cite{elhage2022superposition,netdissect2017,scherlis2022polysemanticity,lecomte2024what,marshall2024understanding,olah2017feature,nguyen2016multifaceted}. This polysemanticity reduces the interpretability of models, motivating the development of Sparse Autoencoders (SAEs) \cite{huben2024sparse,rajamanoharan2024improving,rajamanoharan2024jumping,gao2025scaling}. SAEs are designed to encourage sparsity in neural activations, aiming to align each neuron with a specific, distinct concept and thereby promote monosemanticity and interpretability. In the following, we evaluate SAEs to better understand their effectiveness in improving monosemanticity.

% Although SAEs significantly improve monosemantic behavior, they do not fully eliminate polysemanticity, which continues to persist to some extent.

\begin{figure*}[t]
  \centering
  \includegraphics[width=\textwidth]{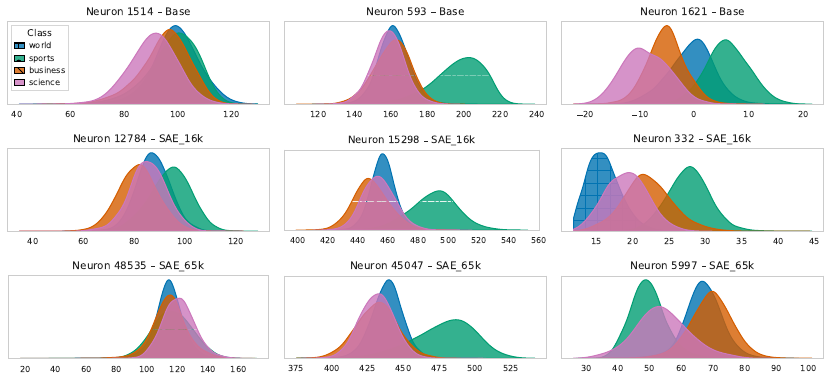}
    \caption{Across base model and SAEs (SAE-16k, SAE-65k), neurons exhibit varying degrees of separability in their activations.
    Some have completely overlapping activations across concepts, others show partial or clear separation.
  This variability underscores the importance of using distribution-aware metrics when assessing monosemanticity.}
  \label{fig:range_patterns_agnews}
\end{figure*}

\subsection{Salient Neuron Overlap}

\vspace{-8pt}

%\begin{figure}[!t]
%  \centering
  %\includegraphics[width=\linewidth]{figures/combined_overlap_plots.pdf}
  %\caption{Salient Neurons Overlap Percentage Across Datasets.}
  %\label{fig:overlap_percentage}
%\end{figure}

%\umaircomment{Add justification for selecting the top activating neurons as related to target concept.} %You can draw motivation from top K SAE's and the general training objective of sparsity in SAEs
%To compute this, we first calculate the mean activation of each neuron for each concept individually. Specifically, for each concept, we average the activations of each neuron across all samples in that concept class. Next, we identify the top-$K$ highly activated neurons per concept by sorting them based on their mean activation values and selecting the top 80. These top-$K$ neurons are assumed to represent the most salient or important neurons for that concept. We then compute two sets: the intersection of top-$K$ neurons across all concepts (i.e., neurons that are highly activated for every concept), and the union of top-$K$ neurons across concepts (i.e., neurons that are salient in at least one concept). The percentage shown in the figure~\ref {fig:overlap_percentage} is the size of the intersection divided by the size of the union. This measures the degree of shared saliency or polysemanticity among concept classes. 

As a first step, we quantify polysemanticity by measuring the overlap percentage of salient neurons (i.e., neurons with high mean activation) across concepts. 
Specifically, for each concept, we identify the top 80 salient neurons by mean activation. Then, the overlap percentage is computed as the intersection-over-union of these top-K sets across concepts. This metric captures the extent to which neurons are shared across concepts, reflecting shared saliency and polysemanticity.
The Figure~\ref{fig:overlap_percentage} compares this shared saliency for the base model and two Sparse Autoencoders (SAEs) with different latent dimensions (16k and 65k), while maintaining comparable sparsity levels (116 vs. 93 active neurons). In the DBpedia dataset, for instance, we observe that nearly 9\% of the top-activated (salient) neurons in the base model are shared across all 14 concepts.
%On the other hand, for the AGNews and POS datasets, this percentage is considerably higher as these datasets contain four concepts, which increases the likelihood that salient neurons are shared across them. 
Moreover, the Figure~\ref{fig:overlap_percentage} confirms that SAEs exhibit reduced conceptual overlap, suggesting less polysemanticity; however, polysemantic neurons are still present. For example, in the POS dataset, the percentage of shared salient neurons drops from around 38\% in the base model to approximately 18\% in the SAE with 16k dimensions. This is still a relatively high percentage, indicating that over 18\% of the most salient neurons are shared across all four concepts. Moreover, another notable observation is that the 65k-dimensional SAEs exhibit lower polysemanticity than their 16k-dimensional counterparts across all datasets. This reinforces the idea that larger SAEs have greater capacity to allocate distinct neurons to specific concepts, thereby enhancing interpretability. For additional analysis evaluating all active SAE neurons, not just the top 80, see Appendix~\ref{app:sae_analysis}.

\subsection{SAEs Activation Distributions}

So far, we have analyzed polysemanticity by quantifying neuron overlap based on activation frequency, specifically, how often a neuron is active or salient across different concepts. While these approaches offer useful aggregate insights, they treat neuron activation as a binary or averaged signal, overlooking the distributional characteristics of how neurons respond to each concept. In other words, a neuron might be shared across multiple concepts, but the manner in which it activates for each could vary significantly, ranging from broad, overlapping responses to distinct, well-separated patterns. To better capture this nuance, Figure~\ref{fig:range_patterns_agnews} displays the full activation distributions of selected neurons across concept classes in the AG News dataset, comparing the base model with two SAE variants of differing capacities. We include an analogous visualization for the NER dataset in Appendix~\ref{appendix:ner_activations}. %\colorbox{yellow}{Qualitative}

These plots reveal two key patterns: (1) consistent with prior observations~\cite{haider2025neurons}, neuron activations in both the base model and SAEs tend to follow approximately Gaussian distributions; and (2) while some neurons are shared across multiple concepts, their activation distributions can range from highly overlapping to clearly separable. This underscores a fundamental limitation of mean-based or binary overlap measurements, which can overlook meaningful distinctions in activation behavior. To more accurately measure polysemanticity, it is therefore necessary to employ a metric that captures the full shape of activation distributions across concepts. To this end, we introduce a new concept separability score, based on the Jensen-Shannon distance, which quantifies the degree of separation between concept-specific activation distributions.

% rather than just central tendency
%Previous work has shown that neuron activations tend to follow an approximately Gaussian distribution \cite{haider2025neurons}. Building on this, we show that neurons in Sparse Autoencoders (SAEs) also exhibit Gaussian-like activation patterns. Figure~\ref{fig:range_patterns} illustrates the activation distributions of salient neurons in layer 25 of the Gemma model (specifically, at the blocks.25.hook\_resid\_post hook, just before the language head) as well as in SAEs with 16k and 65k dimensions at the corresponding post-activation hook. Salient neurons are defined as those with the highest average activation values. All experiments were conducted on the news dataset.

% Motivated by the emergence of Gaussian-like activation patterns in both the base model and SAEs, we propose a new metric to quantify concept separability. This score measures the distance between the Gaussian-like activation distributions corresponding to different classes, providing a more nuanced view of how distinctly concepts are represented in neuron activations.

\begin{table}[t!]
\centering
\resizebox{\columnwidth}{!}{%
\begin{tabular}{l|c|c|c|c|c}
\toprule
& \textbf{POS} & \textbf{AG News} & \textbf{Emotions} & \textbf{DBpedia} & \textbf{NER} \\
\midrule
\textbf{Base}
 & 0.343 & 0.366 & 0.322 & 0.405 & 0.308 \\
\makecell[l]{\textbf{SAE}\\(width\_16k/l0\_116)}
  & 0.539 & 0.650 & 0.431 & 0.621 & 0.581 \\
\makecell[l]{\textbf{SAE}\\(width\_65k/l0\_93)}
  &  \textbf{0.600} & \textbf{0.709} & \textbf{0.446} & \textbf{0.680} & \textbf{0.621} \\
\bottomrule
\end{tabular}%
}
\caption{Separability Score \(S\) across datasets.}
\vspace{-12pt}
\label{tab:JS_separability_score}
\end{table}

\stitle{Concept Separability Score.}  
\label{metric:separability_score}
To quantify how separable a neuron’s activation distributions are across $k$ concepts, we first define the probability density function as:
\[
f_{h^l_j\mid c_i}(x)=p\bigl(h^l_j=x\mid c_i\bigr),\quad i=1,\dots,k,
\]

\noindent with this density function at hand, we define mixture as $M^{(l,j)}(x)=\frac{1}{k}\sum_{i=1}^k f_{h^l_j\mid c_i}(x)$,

\noindent and then compute the generalized Jensen–Shannon divergence \cite{lin1991divergence} as
\[
\resizebox{\columnwidth}{!}{$
\mathrm{JSD}\bigl(f_{h^l_j\mid c_1},\dots,f_{h^l_j\mid c_k}\bigr)
= H\bigl(M^{(l,j)}\bigr)
- \frac{1}{k}\sum_{i=1}^k H\bigl(f_{h^l_j\mid c_i}\bigr).
$}
\]

\noindent Next, we take the square root of the JSD and normalize it to obtain a proper distance metric bounded within $[0,1]$:
\[
\resizebox{\columnwidth}{!}{$
D_{\mathrm{JS}}\bigl(f_{h^l_j\mid c_1},\dots,f_{h^l_j\mid c_k}\bigr)
= \frac{\sqrt{\mathrm{JSD}\bigl(f_{h^l_j\mid c_1},\dots,f_{h^l_j\mid c_k}\bigr)}}{\sqrt{\log_{2}k}},
$}
\]

\noindent Moreover, we assign $D_{\mathrm{JS}}=1$ whenever a neuron’s activations are all attributed to a single concept.  Finally, the layer‐level separability score is
\[
S^l
= \frac{1}{d}\sum_{j=1}^d
D_{\mathrm{JS}}\bigl(f_{h^l_j\mid c_1},\dots,f_{h^l_j\mid c_k}\bigr),
\]
where $d$ is the number of neurons in layer~$l$.

Table~\ref{tab:JS_separability_score} reports the concept separability score \(S \in [0,1]\) for Gemma, where higher values indicate more distinct activation distributions across concepts, indicating higher monosemanticity (Full JS scores for all settings and both LLMs are in Appendix~\ref{app:all_js}). Across all five datasets, SAEs substantially improve separability over the base model. For instance, on DBpedia (14 classes), the score increases from 0.405 in the base model to 0.621 and 0.680 for the 16K- and 65K-dimensional SAEs, respectively, an increase of over 50\%. Furthermore, higher-capacity SAEs consistently yield greater separability, supporting the view that more expressive latent spaces better disentangle concepts.

Building on our finding that sparse autoencoders yield more separable concept representations, we next assess whether this improved separability enables more precise concept erasure interventions.

\section{Concept Erasure}

Concept erasure encompasses interventions that aim to remove a specific concept from a model's internal representation, ideally without affecting other concepts \cite{Dalvi_aaai,Dalvi_Nortonsmith_Bau_Belinkov_Sajjad_Durrani_Glass_2019,dai_etal_2022_knowledge,Ari_2018on}. Formally, let $M$ be a trained model that maps an input $x$ to a concept label $M(x)=c$. An ideal erasure yields a modified model $M'_{\mathrm{ideal}}$ satisfying:
\[
M'_{\mathrm{ideal}}(x) =
\begin{cases}
\neq M(x), & \text{if } M(x)=c, \\
= M(x),   & \text{if } M(x)\neq c.
\end{cases}
\]
That is, the model should unlearn the target concept $c$ while preserving its behavior on all other concepts.

As shown in Figure~\ref{fig:range_patterns_agnews}, some neurons exhibit considerable overlap in their activation distributions across concepts, while others show separability. This pattern appears in both the base model and SAEs, suggesting that concept erasure techniques should not treat all activation values in one neuron identically. Instead, we propose a more targeted approach that considers where an activation falls within the distribution. Specifically, values in regions uniquely tied to a concept (i.e., those regions of distribution that are clearly separable from other concepts) should be suppressed more strongly, while regions shared across concepts should be dampened more conservatively to preserve other concepts. To achieve this, we introduce Attenuation via Posterior Probabilities (\ours), which modulates suppression based on distributional separability.

%As demonstrated in Figure~\ref{fig:range_patterns}, some neurons exhibit considerable overlap in their activation distributions across multiple concepts, while others show strong separability. This pattern holds consistently across the base model and SAEs. These observations suggest that concept erasure techniques should not treat all activation values in one neuron identically. Instead, we are motivated to explore more targeted concept erasure approaches that account for the region where the activation values fall. We propose a more nuanced approach to concept erasure, in which activation values that lie in regions uniquely associated with a specific concept (i.e., those regions of distribution that are clearly separable from other concepts) are suppressed more aggressively. Conversely, activation values in regions shared across multiple concepts should be dampened more conservatively to preserve shared representations and reduce the risk of disrupting unrelated concepts. By modulating the degree of suppression based on distributional separability, we can more precisely remove concept-specific information while minimizing unintended side effects. To do so, we propose Attenuation via Posterior Probabilities (\ours).

\subsection{Attenuation via Posterior Probabilities (\ours)}

Given all neurons \(h^l_j\) of layer \(l\), with individual activations \(x^l_j\) for \(j = 1, \dots, d\), and a target concept \(c_i \in C\), our goal is to selectively suppress the activation \(x^l_j\) that is attributable to \(c_i\), while preserving contributions from other concepts.
We begin by computing the posterior probability that a given activation \(x^l_j\) arose from concept \(c_i\), under the assumption that all concepts are a priori equally likely:

\[
\resizebox{\columnwidth}{!}{$
\begin{aligned}
\pi_{j,i}(x^l_j)
&= p\bigl(c_i \mid h^l_j = x^l_j\bigr)
= \frac{p\bigl(h^l_j = x^l_j \mid c_i\bigr)\,p(c_i)}{\displaystyle\sum_{m=1}^k p\bigl(h^l_j = x^l_j \mid c_m\bigr)\,p(c_m)},\\
&= \frac{p\bigl(h^l_j = x^l_j \mid c_i\bigr)}{\displaystyle\sum_{m=1}^k p\bigl(h^l_j = x^l_j \mid c_m\bigr)}
\;\equiv\;
\frac{f_{h^l_j\mid c_i}(x^l_j)}{\displaystyle\sum_{m=1}^k f_{h^l_j\mid c_m}(x^l_j)}.
\end{aligned}
$}
\]

By definition, \(\sum_{i=1}^{k} \pi_{j,i}(x^l_j) = 1\).

To avoid unreliable posterior estimates from low-density regions, we limit our attention to the central region of the target concept's activation distribution, where density estimates are more reliable. Let \(\mu_{j,i}\) and \(\sigma_{j,i}\) denote the mean and standard deviation of neuron \(h^l_j\) under concept \(c_i\), and define the valid damping window as:

\[
W_{j,i}
\;=\;
\bigl[\mu_{j,i} - 2.5\,\sigma_{j,i},\;\mu_{j,i} + 2.5\,\sigma_{j,i}\bigr].
\]

\noindent The damping factor \(\alpha_{j,i}(x)\) is then defined as:

\[
\alpha_{j,i}(x) =
\begin{cases}
1 - \pi_{j,i}(x), & x \in W_{j,i}, \\
1, & \text{otherwise}.
\end{cases}
\]

\noindent With $\alpha \approx 0$ when $x$ is very typical of the target concept $c_i$. Finally, we apply this factor to dampen the activation:

%\[
%\alpha_{j,i}(x)
%=
%\begin{cases}
%1 \;-\;\displaystyle\frac{f_{h^l_j\mid c_i}(x)}
%                       {\sum_{m=1}^k f_{h^l_j\mid c_m}(x)}, 
%& x \in W_{j,i},\\[1em]
%1, & x \notin W_{j,i}.
%\end{cases}
%\]

\[
\resizebox{\columnwidth}{!}{$
  \widetilde{x}^l_j
    = \alpha_{j,i}(x^l_j)\,x^l_j
    = \begin{cases}
      \bigl[1-\pi_{j,i}(x^l_j)\bigr]\,x^l_j, 
        & |x^l_j-\mu_{j,i}|\le2.5\,\sigma_{j,i},\\[\smallskipamount]
      x^l_j, & \text{otherwise}.
      \end{cases}
$}
\]

This formulation enables precise, concept-aware suppression while leaving unrelated or uncertain activations unchanged.

\subsection{Baseline Methods}
To comprehensively evaluate concept erasure effectiveness, we compare \ours{} (which is a partial suppression method) against three other partial methods and one full-masking baseline.

%We compare our concept erasure approach with the following baselines:  \colorbox{yellow}{Mention partial and full}

\stitle{AURA \cite{aura2024}}: Ranks neurons by AUROC, selects those with $\mathrm{AUROC} > 0.5$, and dampens their output based on  $\mathrm{AUROC}$.

\stitle{Range Masking \cite{haider2025neurons}}: The activations of concept-relevant neurons (highly activated) are suppressed when they fall within their typical range ($\mu \pm 2.5\sigma$).

\stitle{Adaptive Dampening \cite{haider2025neurons}}: The activations of concept-relevant neurons (highly activated) are dampened in proportion to their distance from the concept mean.

\stitle{Full Masking \cite{Dalvi_aaai,dai_etal_2022_knowledge,antverg2022on}}: concept-relevant neurons (highly activated) are fully zeroed out to eliminate the target concept.

%\begin{enumerate}
%  \item \textbf{AURA \cite{aura2024}}: Suau et al. introduce a hyperparameter-free method that ranks neurons by their AUROC for a target concept. Neurons with \(\mathrm{AUROC} > 0.5\) are selected, and their pre-activation outputs \(z_m\) are dampened by a factor \(\alpha_m\), where \(\alpha_m = 0\) for perfect classifiers (AUROC = 1) and \(\alpha_m = 1\) for random ones.

%  \item \textbf{Range Masking \cite{haider2025neurons}}: Haider et al. identify concept-relevant neurons and compute the mean and standard deviation of their activations over concept examples. Activations within the typical range (\(\mu \pm 2.5\sigma\)) are suppressed to remove concept-specific signals while preserving others.

%  \item \textbf{Adaptive Dampening \cite{haider2025neurons}}: A more flexible variant of Range Masking, where suppression strength varies with distance from the concept mean—activations near the mean are dampened more, and those further away are preserved, allowing more precise removal.

%  \item \textbf{Full Masking \cite{Dalvi_aaai,dai_etal_2022_knowledge}}: A simple baseline where all neurons associated with the target concept are completely zeroed out to eliminate the concept from the representation.
%\end{enumerate}

\begin{table*}[!t]
  \centering
  \caption{Concept Erasure Results by Method and Model Type across Datasets (Gemma-2-2B). Bolded values indicate the best performance, and underlined values denote the second-best. Results are grouped by intervention method (e.g., APP, AURA, Adaptive) and model type (Base vs. SAE variants) across five benchmark datasets.}
  \label{tab:gemma_concept_erasure}
  \resizebox{\textwidth}{!}{%
  \begin{tabular}{l l|ccc|ccc|ccc|ccc|ccc}
    \toprule
    \textbf{Type} & \textbf{Method}
      & \multicolumn{3}{c}{\textbf{POS}}
      & \multicolumn{3}{c}{\textbf{AG News}}
      & \multicolumn{3}{c}{\textbf{Emotions}}
      & \multicolumn{3}{c}{\textbf{DBpedia}}
      & \multicolumn{3}{c}{\textbf{NER}} \\
    \cmidrule(lr){3-5} \cmidrule(lr){6-8}
    \cmidrule(lr){9-11} \cmidrule(lr){12-14} \cmidrule(lr){15-17}
    & 
      & \small$\Delta_\mathrm{Acc}\uparrow$ & $\small\Delta_\mathrm{Conf}\uparrow$ &
      \small$\mathrm{DPPL}\downarrow$
      & \small$\Delta_\mathrm{Acc}\uparrow$ & \small$\Delta_\mathrm{Conf}\uparrow$ &
      \small$\mathrm{DPPL}\downarrow$
      & \small$\Delta_\mathrm{Acc}\uparrow$ & \small$\Delta_\mathrm{Conf}\uparrow$ &
      \small$\mathrm{DPPL}\downarrow$
      & \small$\Delta_\mathrm{Acc}\uparrow$ & \small$\Delta_\mathrm{Conf}\uparrow$ &
      \small$\mathrm{DPPL}\downarrow$
      & \small$\Delta_\mathrm{Acc}\uparrow$ & \small$\Delta_\mathrm{Conf}\uparrow$ &
      \small$\mathrm{DPPL}\downarrow$\\
    \midrule
    \multirow{5}{*}{Base}
      & \ours & \textbf{0.224} & \textbf{0.210} & \textbf{0.571}
                    & \textbf{0.347} & \textbf{0.266} & \textbf{0.394}
                    & \textbf{0.051} & 0.095 & \textbf{0.151}
                    & 0.113 & \textbf{0.104}  & \textbf{0.076}
                    & 0.079 & 0.129 & \textbf{0.201}\\
      & Aura         & 0.063 & 0.091 & 1.195
                    & \underline{0.112} &  0.092 & 2.130
                    & -0.019 &  0.004 & 2.273
                    & 0.092 & 0.064  & 2.792
                    & 0.033 & 0.044 & 1.515\\ 
      & Range        & 0.151 & 0.1006 & 2.258
                    & 0.081 &  \underline{0.098} & 1.026
                    & 0.036 &  \underline{0.104} & 0.784
                    & \textbf{0.168} & \textbf{0.109}  & 0.700
                    & \textbf{0.135} & \textbf{0.209} & 1.978\\ 
      & Adaptive     & \underline{0.157} & \underline{0.129} & \underline{1.158}
                    & 0.050 & 0.089 & \underline{0.545}
                    & 0.032 &  0.090 & \underline{0.428}
                    & \underline{0.120} & 0.099  & \underline{0.387}
                    & \underline{0.095} & 0.173 & \underline{1.052}\\ 
      & Full         & 0.115 & 0.069 & 22.873
                    & 0.096 &  0.080 & 16.914
                    & \underline{0.038} & \textbf{0.113} & 17.438
                    & \textbf{0.169} & \underline{0.102}  & 10.922 
                    & \textbf{0.134} & \underline{0.176} & 23.598\\
    \midrule
    \multirow{5}{*}{\shortstack{SAE\\width: 65k\\l0: 93}}
      & \ours & \underline{0.302} & \underline{0.270} & \textbf{0.342}
                    & \textbf{0.601} & \textbf{0.442}  & \textbf{0.230}
                    & \underline{0.231} & \underline{0.279}  & \textbf{0.075}
                    & \textbf{0.399} & \textbf{0.219}  & \textbf{0.114}
                    & \underline{0.507} & 0.520 & \textbf{0.161} \\ 
      & Aura         & \textbf{0.406} & \textbf{0.312}  & \underline{0.360}
                    & 0.577 &  \underline{0.367} & 0.592
                    & \textbf{0.336} &  \textbf{0.311} & 0.286
                    & \underline{0.357} & \textbf{0.213}  & 0.490
                    & \textbf{0.519} & 0.514  & 0.359 \\
      & Range        & 0.052 & 0.055 & 0.741
                    & 0.587 & 0.125  & 0.542
                    & 0.146 & 0.024  &  0.413
                    & 0.252 & 0.056  & 0.446
                    & 0.497 & \underline{0.549}  & 0.433 \\ 
      & Adaptive     & 0.095 & 0.105  & 0.533
                    &  \underline{0.592} &  0.215 & \underline{0.358}
                    & 0.182 & 0.084  & \underline{0.226}
                    & 0.222 & \underline{0.083} & \underline{0.279}
                    & 0.494 & \textbf{0.557} & \underline{0.308}\\ 
      & Full         & 0 & 0.0003 & 1.712
                    & \textbf{0.602} & 0.006 & 4.763
                    & 0.120 & 0.002 & 5.462
                    & 0.275 & 0.018 & 3.892
                    & 0.382 & 0.391 & 3.046\\
    \midrule
    \multirow{5}{*}{\shortstack{SAE\\width: 16k\\l0: 116}}
      & \ours & \underline{0.216} & \textbf{0.191}  & \textbf{0.468}
                    & \underline{0.582} & \textbf{0.369} & \textbf{0.569}
                    & \underline{0.265} & \underline{0.274} & \textbf{0.166}
                    & \textbf{0.312} & \textbf{0.152} & \textbf{0.140}
                    & \underline{0.377} & \underline{0.397} & \textbf{0.531} \\ 
      & Aura         & \textbf{0.250} & \textbf{0.192} & \underline{0.577}
                    & \textbf{0.596} & \underline{0.286} & 1.262
                    & \textbf{0.339} & \textbf{0.280} & 0.493
                    & \underline{0.257} & \underline{0.140} & 0.836
                    & \textbf{0.389} & \textbf{0.415} & 0.741 \\ 
      & Range        & 0.061 & 0.046 & 0.925
                    & 0.505 & 0.103 & 0.879
                    & 0.099 & 0.009 & 0.418
                    & 0.208 & 0.053 & 0.446
                    & 0.298 & 0.305 & 0.915 \\ 
      & Adaptive     & 0.097 & \underline{0.085} & 0.655
                    & 0.508 & 0.163 & \underline{0.676}
                    & 0.128 & 0.047 & \underline{0.267}
                    & 0.177 & 0.071 & \underline{0.288} 
                    & 0.329 & 0.347 & \underline{0.721} \\ 
      & Full         & 0 & 0.0005 & 3.738
                    & 0.462 & 0.001 & 5.882
                    & 0.048 & 0.0005 & 6.454
                    & 0.187 & 0.0007 & 5.639
                    & 0.262 & 0.256 & 3.872\\
    \midrule
    \multirow{5}{*}{\shortstack{SAE\\width: 65k\\l0: 197}}
      & \ours & \underline{0.350} & \textbf{0.322} & \textbf{0.398}
                    & \textbf{0.615} & \textbf{0.458} & \textbf{0.433}
                    & \underline{0.247} & \underline{0.285} & \textbf{0.406}
                    & \textbf{0.362} & \textbf{0.235} & \textbf{0.083}
                    & \underline{0.530} & 0.654 & \textbf{0.346}\\ 
      & Aura         & \textbf{0.384} & \underline{0.317} & \underline{0.576}
                    & \textbf{0.611} & \underline{0.397} & 0.940
                    & \textbf{0.311} & \textbf{0.327} & 0.584
                    & \underline{0.297} & \textbf{0.235} & 0.596
                    & \textbf{0.567} & \textbf{0.681} & \underline{0.510} \\
      & Range        & 0.046 & 0.059 & 0.902
                    & \underline{0.346} & 0.060 & 0.748
                    & 0.135 & 0.024 & 0.729
                    & 0.292 & 0.106 & 0.335
                    & 0.503 & \underline{0.662} & 0.613 \\ 
      & Adaptive     & 0.110 & 0.130 & 0.634
                    & 0.345 & 0.130 & \underline{0.554}
                    & 0.170 & 0.110 & \underline{0.574}
                    & 0.257 & \underline{0.147} & \underline{0.205}
                    & 0.503 & 0.656 & 0.815 \\ 
      & Full         & 0 & 0.00001 & 5.599
                    & 0.344 & 0.0007 & 7.422
                    & 0.110 & 0.002 & 7.613
                    & 0.216 & 0.019 & 6.203 
                    & 0.292 & 0.384 & 3.625 \\
    \midrule
    \multirow{5}{*}{\shortstack{SAE\\width: 16k\\l0: 285}}
      & \ours & \underline{0.382} & \underline{0.306} & \textbf{0.634}
                    & \textbf{0.628} & \textbf{0.356} & \textbf{0.882}
                    & \underline{0.212} & \underline{0.228} & \textbf{0.492}
                    & \textbf{0.274} & \textbf{0.147} & \textbf{0.167}
                    & \textbf{0.423} & \textbf{0.482}  & \textbf{0.370}\\ 
      & Aura         & \textbf{0.433} & \textbf{0.359} & \underline{0.778}
                    & \underline{0.549} & \underline{0.346} & 1.671
                    & \textbf{0.286} & \textbf{0.300} & \underline{0.788}
                    & 0.158 & \underline{0.111} & 0.975
                    & \textbf{0.428} & \underline{0.469}  & 0.718\\ 
      & Range        & 0.051 & 0.047 & 1.356
                    & 0.456 & 0.069 & 1.852
                    & 0.111 & 0.012 &  1.395
                    & \underline{0.181} & 0.057 & 0.958 
                    & 0.338 &  0.392 & 0.924\\ 
      & Adaptive     & 0.179 & 0.148 & 0.975
                    & 0.441 & 0.149 & \underline{1.312}
                    & 0.121 & 0.072 & 0.948
                    & 0.145 & 0.083 & \underline{0.582}
                    & \underline{0.376} & 0.435  & \underline{0.669}\\ 
      & Full         & 0 & 0.001 & 5.880
                    & 0.213 & 0.0004 & 16.284
                    & 0.062 & 0.001 &  8.117
                    & 0.153 & 0.023 & 9.510
                    & 0.167 & 0.191 & 8.661\\
    \bottomrule
  \end{tabular}
  }
  \vspace{-10pt}
\end{table*}

\subsection{Metrics}

We evaluate the causal effect of our interventions using three metrics: task accuracy, predictive confidence, and perplexity.

Accuracy and confidence are measured both before and after intervention, for the target concept \(c\) and all auxiliary concepts \(c' \neq c\). The goal is to assess how much the intervention selectively affects the target concept while minimizing disruption to others.

Let \( D_{\mathrm{Acc}} \) denote the drop in accuracy for the target concept, and \( D'_{\mathrm{Acc}} \) the average drop in accuracy across auxiliary concepts. Similarly, let \( D_{\mathrm{Conf}} \) and \( D'_{\mathrm{Conf}} \) be the drops in predictive confidence for the target and auxiliary concepts, respectively.

Using these, we compute two scores:
\[
\Delta_{\mathrm{Acc}} = D_{\mathrm{Acc}} - D'_{\mathrm{Acc}}, \quad
\Delta_{\mathrm{Conf}} = D_{\mathrm{Conf}} - D'_{\mathrm{Conf}}.
\]

Higher values of \( \Delta_{\mathrm{Acc}} \) and \( \Delta_{\mathrm{Conf}} \) indicate more precise interventions, strongly affecting the target concept while preserving performance on others. In the main text, we report only \( \Delta_{\mathrm{Acc}} \) and \( \Delta_{\mathrm{Conf}} \); the full metric breakdowns are included in Appendix~\ref{app:details_metrics}.

Lastly, to capture the overall impact on the model’s generative ability, we measure the increase in perplexity:
\[
\mathrm{DPPL} = \mathrm{PPL}_{\mathrm{post}} - \mathrm{PPL}_{\mathrm{base}} .
\]

Comprehensive implementation details are presented in Appendix~\ref{appendix:implementation_details}, 
including our histogram-based KDE for concept-conditioned densities.

\subsection{Results and Analysis}

\stitle{Comparison of Intervention Effectiveness: SAEs vs. Base Model.}
As it can be seen in Table~\ref{tab:gemma_concept_erasure}, across nearly all settings, we find that partial intervention methods (particularly {\ours} and AURA) consistently achieve higher $\Delta_{\mathrm{Acc}}$ and $\Delta_{\mathrm{Conf}}$ when applied to SAE representations compared to the base model. Specifically, for {\ours}, SAE-based interventions outperformed the base model in 38 out of 40 comparisons. For AURA, SAE-based interventions were more effective in all 40 cases. In contrast, full masking shows less benefit from SAE representations; 19 out of 40 interventions resulted in better outcomes than when applied to the base model. This discrepancy suggests that coarse suppression methods fail to capitalize on the increased concept separability offered by SAEs. These findings reinforce that SAE representations are more disentangled and that fine-grained, distribution-aware methods are better equipped to exploit this structure for effective concept removal. Furthermore, focusing specifically on the {\ours} intervention (which is a partial intervention), we observe that within the SAE family, increasing capacity consistently enhances intervention quality. In particular, $\Delta_{\mathrm{Conf}}$ consistently increases as we scale from 16\,k to 65\,k latent dimensions, reflecting improved confidence suppression for the target concept. $\Delta_{\mathrm{Acc}}$ also improves across most datasets, with only minor exceptions, further underscoring the role of latent dimensionality in enabling more precise and effective concept removal.

\stitle{Partial Interventions Vs Full Interventions.}
Full masking ranks as the worst‐performing method in 36 out of 40 $\Delta$-metrics on the SAEs and it also produces the largest perplexity increase in all DPPL evaluations across SAEs and the base model. This underscores that distribution-aware partial methods (e.g., {\ours}, AURA), which leverage activation distributions, are far more effective for targeted concept removal than the coarse, distribution-agnostic full-masking approach. %\colorbox{yellow}{need work}
%and in 1 out of 10 $\Delta$-metrics on the base model

\stitle{{\ours} is least disruptive and highly competitive on concept erasure.}
{\ours} consistently achieves the smallest $\mathrm{DPPL}$ across all 25 experiments, making it the least disruptive method. Adaptive follows as the second-best approach, attaining the second-smallest $\mathrm{DPPL}$ in 19 out of 25 cases. For concept removal, {\ours} demonstrates strong and consistent performance; it most frequently achieves the best $\Delta_{Conf}$ (14/25 cases), while ranking first or second in 21 out of 25 settings. Regarding $\Delta_{Acc}$, AURA leads with 14 wins, though {\ours} follows closely and ranks in the top two positions for 23 out of 25 settings. Overall, {\ours} effectively removes target concepts while preserving predictive fluency better than all other baselines.

\stitle{Comparative Analysis of {\ours} and AURA.} 
\label{title:aura_and_ours_analysis}
The superior effectiveness of AURA and \ours\ on both $\Delta_{\mathrm{Acc}}$ and $\Delta_{\mathrm{Conf}}$ stems from the fact that they explicitly model not only the target‐concept distribution but also the distributions of all auxiliary concepts. By calibrating their interventions to maximize disruption of $c$ while minimizing collateral effects on $c' \neq c$, both methods achieve higher $\Delta$ values than approaches that consider only the target distribution. Between these two, \ours\ pulls ahead of AURA in terms of perplexity ($\mathrm{DPPL}$) because it leverages fine‐grained, activation‐specific damping rather than a single, per‐neuron factor. AURA mutes an “expert” neuron uniformly, regardless of whether a particular activation is highly characteristic of the target concept, whereas \ours\ computes $\pi_{j,i}(x)$ for each activation and suppresses only the portions of the distribution uniquely associated with $c_i$. This activation-aware attenuation not only removes the targeted concepts effectively but also best preserves the model’s overall fluency, as evidenced by the smallest $\mathrm{DPPL}$. %among all baselines.

 %However, it performs less favorably on the IMDB and SST2 datasets when applied to the Base model. Because these datasets contain only two concepts, the method focuses on preserving one while removing the other, ignoring the broader conceptual space. This narrow preservation can inadvertently affect general representations, leading to increased perplexity.

\stitle{Cross-Model Validation (DeepSeek).}
To verify that our observations are not specific to Gemma-2, we replicate experiments on DeepSeek-R1 (Appendix~\ref{appendix:deepseek_concept_erasure}). The results show the same trends, SAEs enable more selective concept removal than the base model. Moreover, \ours{} achieves the smallest perplexity degradation ($\mathrm{DPPL}$) across all datasets except one, while remaining highly competitive on the concept-removal metrics ($\Delta_{\mathrm{Acc}}$, $\Delta_{\mathrm{Conf}}$).
%SAEs increase JS-based separability and enable more targeted concept removal than the base model.

%\subsection{Relation Between Concept Erasure Methods and JS Distance}
\subsection{Relation Between Erasure Methods and JS Distance}

\begin{figure}[t]
    \centering
    \includegraphics[width=\columnwidth]{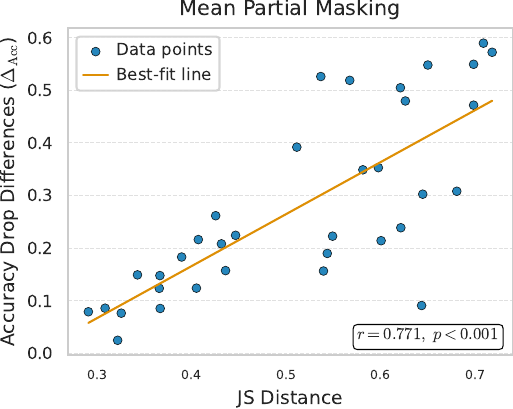}
    \caption{Separability Score vs. Erasure Ability (Partial)}
    \label{fig:partial_vs_js}
    \vspace{-8pt}
\end{figure}

\begin{figure}[t]
    \centering
    \includegraphics[width=\columnwidth]{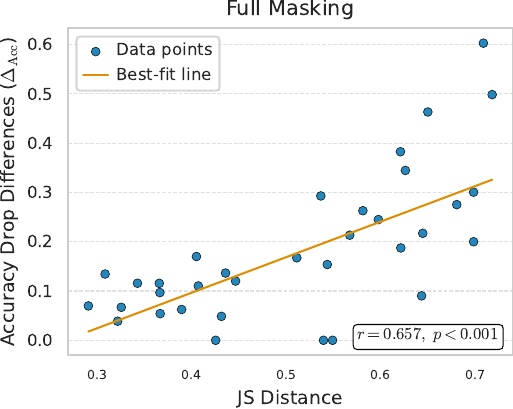}
    \caption{Separability Score vs. Erasure Ability (Full)}
    \label{fig:full_vs_js}
    \vspace{-8pt}
\end{figure}

Our distribution-aware separability score $S$ (see Subsection~\ref{metric:separability_score}) serves as a natural predictor for the precision of concept erasure methods. Intuitively, the more separable a neuron’s activation distributions are, the easier it should be to suppress only the target concept while preserving unrelated behavior.
We empirically validate this hypothesis in Figure~\ref{fig:partial_vs_js}, which plots our JS separability score $S$ (x-axis) against the change in accuracy difference $\Delta\text{Acc}$ (y-axis), aggregating data from both the Gemma and DeepSeek experiments (Both SAE and base models). A strong positive Pearson correlation ($r = 0.771$, p < 0.001) emerges for the average performance of partial erasure methods (\ours{}, AURA, Range, and Adaptive), confirming that higher separability reliably predicts more selective accuracy drops on the target concept. Detailed per-method correlations are provided in Appendix \ref{app:correlations}.
For comparison, we conducted the same analysis for the full masking approach (Figure~\ref{fig:full_vs_js}). While a positive correlation also appears ($r = 0.657$, p < 0.001), it is notably weaker than that observed for partial masking.
This gap reinforces a key insight that full masking cannot exploit the fine-grained separability of activation distributions, whereas distribution-aware partial methods (e.g., APP, AURA) do.%; together with the performance gaps in Section 4.4 (Table 2), these correlations (Figs. 3–4) provide converging evidence for this claim. 
%—yielding more precise and minimally invasive concept erasure
%This gap underscores a key insight which is full masking fails to exploit the finer-grained separability of activation distributions, while distribution-aware partial methods (e.g., APP, AURA) can more effectively capitalize on this structure to achieve precise and minimally invasive concept erasure.

%Figures~\ref{fig:app_vs_js} and~\ref{fig:aura_vs_js} empirically confirm this connection. Each plot relates the JS-based separability score $S$ (x-axis) to the change in accuracy differences $\Delta\text{Acc}$ (y-axis), aggregating points across tasks and settings. For {\ours}, . For AURA, the same relationship holds with a slightly weaker but still strong $r = 0.719$ (\textit{p}~$<$~0.001). 
%Taken together, Figures~\ref{fig:app_vs_js}--\ref{fig:aura_vs_js} support the central claim that JS-distance separability is a practical indicator of erasure efficacy, and among partial, distribution-aware methods, APP exhibits the strongest concordance with this indicator. 
%Additional correlation analyses (e.g., $S$ vs.~$\Delta\text{Conf}$, $S$ vs.~$\Delta\text{PPL}$, and per-dataset/per-layer breakdowns) are provided in the Appendix.

\section{Related Works}

\subsection{Sparse Autoencoders for Feature Discovery} Sparse Autoencoders (SAEs) have emerged as a powerful method for learning interpretable, monosemantic features from neural network activations \cite{huben2024sparse}. Recent advances have focused on improving reconstruction quality and scaling through architectural and training strategy innovations such as JumpReLU activations, BatchTopK sparsity, gated, and end-to-end training frameworks \cite{rajamanoharan2024improving, rajamanoharan2024jumping,gao2025scaling,bussmann2024batchtopk,braun2024identifying}.
Empirical analyses have validated SAEs' ability to discover meaningful structures across different domains, from vision-language models to algorithmic patterns like temporal difference learning in LLMs \cite{sun2025dense,pach2025sparse,demircan2025sparse}. Moreover, evaluation studies have highlighted both their utility for interpretability tasks and remaining challenges with polysemantic representations \cite{kantamneni2025are,minegishi2025rethinking,karvonen2025saebench}. \emph{However, none of these empirical analyses evaluated the separability of activation distributions in SAEs as a measure of polysemanticity.}

\subsection{SAE-Based Model Control} The interpretable features learned by SAEs enable precise control over language model behavior. Several works have demonstrated effective steering by carefully selecting and manipulating SAE features, with approaches ranging from supervised methods for identifying relevant dimensions to frameworks using hypernetworks \cite{arad2025saes,he2025sae,he2025saif,bayat2025steering, sun2025hypersteer,minegishi2025rethinking}. \emph{However, prior SAE-based control methods did not utilize posterior probabilities, limiting their precision}. %Recent work has also explored the appropriate scope of SAE-based interventions, suggesting they are better suited for discovering unknown concepts rather than manipulating known ones \cite{peng2025use}.

\subsection{Base Model Control and Causal Analysis}
Complementing SAE-based approaches, researchers have developed techniques for direct activation control and causal analysis in the base language models. General frameworks for transporting activations facilitate intervention across model architectures \cite{rodriguez2025controlling}, while causal tracing methods enable precise localization and editing of specific knowledge or biases \cite{NEURIPS2020_92650b2e,meng2022locating,meng2023massediting}. These approaches offer foundational tools for probing and manipulating model behavior at the activation level. \emph{However, again, they do not leverage posterior probabilities for better intervention and have been applied exclusively to base models, not to representations learned by SAEs}.

%Complementing SAE-based approaches, researchers have developed methods for direct activation control and causal analysis of language models. General frameworks for transporting activations enable control across different architectures \cite{rodriguez2025controlling}, while causal tracing techniques allow precise localization and editing of specific knowledge and biases \cite{NEURIPS2020_92650b2e,meng2022locating,meng2023massediting}. These methods provide foundational tools for understanding and manipulating model behavior at the activation level. Similarly, these methods did not tak e advntage of posterior probabilities and they were only applied to base model and not SAEs.

%1- Talk about the research that has been done in SAEs 
%SAEs main papers: \cite{huben2024sparse}, \cite{rajamanoharan2024improving}, \cite{rajamanoharan2024jumping}, \cite{gao2025scaling}, \cite{bussmann2024batchtopk}, \cite{braun2024identifying}

%SAEs analysis papers:
%\cite{sun2025dense} \cite{kantamneni2025are} \cite{pach2025sparse} \cite{demircan2025sparse} \cite{minegishi2025rethinking}

%2- controlling with SAEs \cite{arad2025saes}, \cite{he2025sae}, \cite{he2025saif}, \cite{bayat2025steering}, \cite{peng2025use}?, \cite{sun2025hypersteer}

%3- controlling on base model 
%\cite{rodriguez2025controlling}

%Causal tracing: \cite{NEURIPS2020_92650b2e}, \cite{meng2022locating} \cite{meng2023massediting}

\section{Conclusion}
This work presents the first quantitative analysis of monosemanticity in SAEs compared to their dense base models through distributional lens.
%We show that while SAEs reduce polysemanticity and improve concept separability, increased sparsity does not always enhance interpretability and often comes with trade-offs in downstream performance. 
To better characterize monosemanticity, we introduce an activation distribution-aware concept separability score based on the Jensen–Shannon distance, which captures fine-grained distinctions in neuron activations across concepts.
We also demonstrate that SAEs support more precise concept-level interventions than base models, particularly when using partial suppression. Building on this, we propose a new method, Attenuation via Posterior Probabilities, which achieves effective concept removal with least possible side effects. %Together, these findings highlight the importance of fine-grained evaluation metrics and suggest that sparsity, when appropriately applied, can offer both interpretability and control in large language models.

\section{limitations}

To make density estimation computationally feasible at scale, \ours{} replaces standard kernel density estimation (KDE) with a histogram-based approximation. While this approach substantially improves efficiency, it also introduces certain limitations. As the number of histogram bins increases, the accuracy of the estimated activation distributions improves, but so does the computational cost. Consequently, achieving the best possible performance of \ours{}, in terms of precise density estimation and separability, requires significantly higher computational resources and runtime. Future work could explore alternative KDE methods to better balance accuracy and efficiency.

%While our framework \ours{} could, in principle, be applied to concept erasure in domains such as toxicity mitigation, this direction was not the primary focus of our study. In particular, we did not conduct experiments on toxicity datasets such as those used by AURA, which is specifically designed for that task. Our emphasis was on measuring concept separability and interpretability rather than on task-specific erasure. Exploring the applicability of \ours{} to toxicity removal settings remains an important direction for future work.

\bibliography{custom}

\appendix

\appendix

\setcounter{secnumdepth}{1}

%\clearpage

\vspace{1em}  % Space between figure and section title

\begin{figure*}[!t]
  \centering
  \includegraphics[width=\textwidth]{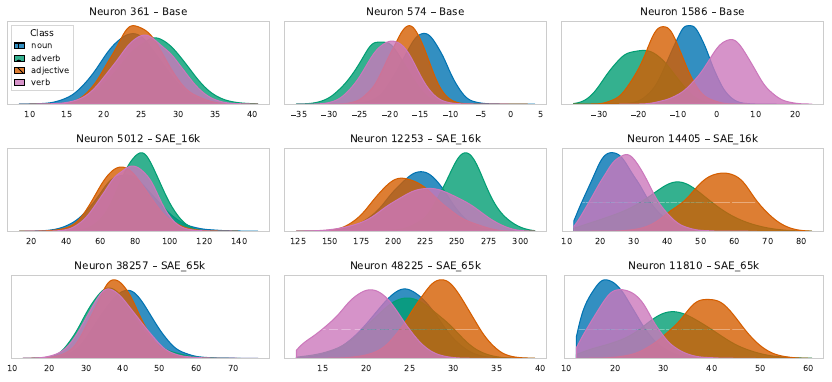}
  \caption{Across base model and SAEs (SAE-16k, SAE-65k), neurons exhibit varying degrees of separability in their activations.
  Some have completely overlapping activations across concepts, others show partial or clear separation.
  This variability underscores the importance of using distribution-aware metrics when assessing neuron monosemanticity.}
  \vspace{-15pt}
  \label{fig:range_patterns_ner}
\end{figure*}

\begin{table*}[t]
  \centering
  \caption{Concept Erasure Results (Deepseek)}
  \label{tab:deepseek_concept_erasure}
  \resizebox{\textwidth}{!}{%
  \begin{tabular}{l l|ccc|ccc|ccc|ccc|ccc}
    \toprule
    \textbf{Type} & \textbf{Method}
      & \multicolumn{3}{c}{\textbf{POS}}
      & \multicolumn{3}{c}{\textbf{AG News}}
      & \multicolumn{3}{c}{\textbf{Emotions}}
      & \multicolumn{3}{c}{\textbf{DBpedia}}
      & \multicolumn{3}{c}{\textbf{NER}} \\
    \cmidrule(lr){3-5} \cmidrule(lr){6-8}
    \cmidrule(lr){9-11} \cmidrule(lr){12-14} \cmidrule(lr){15-17}
    & 
      & \small$\Delta_\mathrm{Acc}\uparrow$ & \small$\Delta_\mathrm{Conf}\uparrow$ & \small$\mathrm{DPPL}\downarrow$
      & \small$\Delta_\mathrm{Acc}\uparrow$ & \small$\Delta_\mathrm{Conf}\uparrow$ & \small$\mathrm{DPPL}\downarrow$
      & \small$\Delta_\mathrm{Acc}\uparrow$ & \small$\Delta_\mathrm{Conf}\uparrow$ & \small$\mathrm{DPPL}\downarrow$
      & \small$\Delta_\mathrm{Acc}\uparrow$ & \small$\Delta_\mathrm{Conf}\uparrow$ & \small$\mathrm{DPPL}\downarrow$
      & \small$\Delta_\mathrm{Acc}\uparrow$ & \small$\Delta_\mathrm{Conf}\uparrow$ & \small$\mathrm{DPPL}\downarrow$\\
    \midrule
    \multirow{5}{*}{Base}
      & \ours & \textbf{0.185} & \underline{0.210} & \underline{0.269}
                    & \textbf{0.184} & \textbf{0.384} & \textbf{0.182}
                    & \textbf{0.100} & \textbf{0.191} & \textbf{0.031}
                    & \underline{0.155} & 0.251 & \textbf{0.133}
                    & \underline{0.089} & 0.143 & \textbf{0.043}\\
      & Aura         & 0.049 & 0.069 & \textbf{0.253}
                    & 0.015 & 0.071 & \underline{0.250}
                    & 0.067 & 0.146 & \underline{0.205}
                    & \textbf{0.211} & 0.281 & \underline{0.607}
                    & 0.038 & 0.071 & \underline{0.284}\\ 
      & Range        & \underline{0.057} & \textbf{0.223} & 0.722
                    & \underline{0.174} & 0.217 & 1.100
                    & \underline{0.078} & 0.151 & 0.427
                    & 0.149 & \textbf{0.325} & 1.191
                    & \underline{0.089} & \textbf{0.256} & 1.315\\ 
      & Adaptive     & 0.050 & 0.139 & 0.354
                    & 0.122 & \underline{0.276} & 0.491
                    & 0.072 & \underline{0.160} & 0.249
                    & 0.115 & 0.244 &  0.752
                    & \textbf{0.090} & \underline{0.198} & 0.583\\ 
      & Full         & 0.054 & 0.181 & 2.710
                    & 0.115 & 0.115 & 5.665
                    & 0.070 & 0.135 & 1.403
                    & 0.136 & \underline{0.289} &  4.710
                    & 0.067 & 0.173 & 6.895\\
    \midrule
    \multirow{5}{*}{SAE}
      & \ours & 0.307 & 0.327 & \underline{7.077}
                    & \underline{0.487} & 0.809 & \textbf{4.244}
                    & \underline{0.092} & 0.473 & \textbf{1.320}
                    & \textbf{0.651} & 0.658 & \textbf{1.515}
                    & \underline{0.567} & \underline{0.877} & \textbf{3.393}\\ 
      & Aura         & \textbf{0.414} & 0.412 & \textbf{2.461}
                    & \textbf{0.514} & 0.774 & \underline{4.450}
                    & 0.089 & 0.410 & \underline{1.439}
                    & \textbf{0.652} & \textbf{0.774} & 6.279
                    & \textbf{0.589} & \textbf{0.883} & \underline{4.206}\\
      & Range        & \underline{0.353} & \underline{0.422} & 15.374
                    & 0.453 & \textbf{0.846} & 8.247
                    & 0.090 & \underline{0.658} & 1.926
                    & \underline{0.498} & 0.646 &  2.715
                    & 0.540 & 0.830 & 11.409 \\ 
      & Adaptive     & 0.338 & \textbf{0.431} & 12.211
                    & 0.433 & \underline{0.831} & 6.934
                    & \textbf{0.093} & 0.652 & 1.538
                    & 0.488 & 0.631 & \underline{2.287}
                    & 0.501 & 0.789 & 9.057\\ 
      & Full         & 0.245 & 0.307 & 23.146
                    & 0.200 & 0.478 & 30.988
                    & 0.090 & \textbf{0.744} & 3.085
                    & \underline{0.498}  & \underline{0.676} &  6.877
                    & 0.300 & 0.489 & 29.032\\
    \bottomrule
  \end{tabular}
  }
\end{table*}

\FloatBarrier

%\clearpage

\begin{table*}[t!]
  \centering
  \caption{Concept Erasure Detailed metrics \(D_{\mathrm{Acc}}\), \(D'_{\mathrm{Acc}}\), \(D_{\mathrm{Conf}}\), and \(D'_{\mathrm{Conf}}\) for Gemma-2-2b.}
  \label{tab:gemma_concept_erasure_details}
  \resizebox{\textwidth}{!}{%
  \begin{tabular}{l l
                  |cccc % POS: was ccc
                  |cccc % AG News
                  |cccc % Emotions
                  |cccc % DBpedia
                  |cccc % NER
                  }
    \toprule
    \textbf{Type} & \textbf{Method}
      & \multicolumn{4}{c}{\textbf{POS}}
      & \multicolumn{4}{c}{\textbf{AG News}}
      & \multicolumn{4}{c}{\textbf{Emotions}}
      & \multicolumn{4}{c}{\textbf{DBpedia}}
      & \multicolumn{4}{c}{\textbf{NER}} \\
    % adjust these spans to 4 columns each:
    \cmidrule(lr){3-6} \cmidrule(lr){7-10}
    \cmidrule(lr){11-14} \cmidrule(lr){15-18} \cmidrule(lr){19-22}
    & 
      & \small$D_{\mathrm{Acc}}\uparrow$
      & \small$D'_{\mathrm{Acc}}\downarrow$
      & \small$D_{\mathrm{Conf}}\uparrow$
      & \small$D'_{\mathrm{Conf}}\downarrow$
      & \small$D_{\mathrm{Acc}}\uparrow$
      & \small$D'_{\mathrm{Acc}}\downarrow$
      & \small$D_{\mathrm{Conf}}\uparrow$
      & \small$D'_{\mathrm{Conf}}\downarrow$
      & \small$D_{\mathrm{Acc}}\uparrow$
      & \small$D'_{\mathrm{Acc}}\downarrow$
      & \small$D_{\mathrm{Conf}}\uparrow$
      & \small$D'_{\mathrm{Conf}}\downarrow$
      & \small$D_{\mathrm{Acc}}\uparrow$
      & \small$D'_{\mathrm{Acc}}\downarrow$
      & \small$D_{\mathrm{Conf}}\uparrow$
      & \small$D'_{\mathrm{Conf}}\downarrow$
      & \small$D_{\mathrm{Acc}}\uparrow$
      & \small$D'_{\mathrm{Acc}}\downarrow$
      & \small$D_{\mathrm{Conf}}\uparrow$
      & \small$D'_{\mathrm{Conf}}\downarrow$ \\
    \midrule 
    \multirow{5}{*}{Base} & \ours & 0.276 & 0.051 & 0.209 & -0.002 & 0.356 & 0.008 & 0.254 & -0.013 & 0.066 & 0.015 & 0.090 & -0.005 & 0.126 & 0.013 & 0.101 & -0.003 & 0.092 & 0.012 & 0.124 & -0.006 \\
    & Aura & 0.128 & 0.065 & 0.131 & 0.039 & 0.151 & 0.039 & 0.176 & 0.084 & 0.027 & 0.047 & 0.033 & 0.029 & 0.170 & 0.077 & 0.095 & 0.031 & 0.069 & 0.036 & 0.093 & 0.048 \\ 
    & Range & 0.699 & 0.547 & 0.546 & 0.446 & 0.276 & 0.194 & 0.328 & 0.229 & 0.125 & 0.089 & 0.210 & 0.105 & 0.357 & 0.189 & 0.195 & 0.085 & 0.259 & 0.124 & 0.362 & 0.152 \\
    & Adaptive & 0.567 & 0.409 & 0.446 & 0.317 & 0.145 & 0.094 & 0.220 & 0.131 & 0.096 & 0.064 & 0.148 & 0.058 & 0.239 & 0.118 & 0.136 & 0.037 & 0.176 & 0.081 & 0.237 & 0.063 \\ 
    & Full & 0.710 & 0.594 & 0.553 & 0.483 & 0.284 & 0.187 & 0.328 & 0.248 & 0.126 & 0.087 & 0.212 & 0.099 & 0.364 & 0.194 & 0.196 & 0.093 & 0.259 & 0.125 & 0.358 & 0.182 \\ 
    \midrule 
    \multirow{5}{*}{\shortstack{SAE\\width: 65k\\l0: 93}}
    & \ours & 0.581 & 0.278 & 0.386 & 0.116 & 0.624 & 0.022 & 0.412 & -0.031 & 0.268 & 0.036 & 0.278 & -0.001 & 0.411 & 0.012 & 0.216 & -0.004 & 0.541 & 0.033 & 0.548 & 0.028 \\ 
    & Aura & 0.615 & 0.209 & 0.388 & 0.075 & 0.593 & 0.015 & 0.411 & 0.044 & 0.374 & 0.037 & 0.318 & 0.007 & 0.413 & 0.055 & 0.216 & 0.003 & 0.554 & 0.034 & 0.560 & 0.045 \\
    & Range & 0.676 & 0.624 & 0.494 & 0.438 & 0.672 & 0.085 & 0.434 & 0.309 & 0.414 & 0.268 & 0.381 & 0.357 & 0.412 & 0.159 & 0.219 & 0.163 & 0.552 & 0.054 & 0.557 & 0.007 \\ 
    & Adaptive & 0.675 & 0.579 & 0.492 & 0.386 & 0.669 & 0.077 & 0.433 & 0.217 & 0.413 & 0.230 & 0.378 & 0.294 & 0.411 & 0.189 & 0.218 & 0.134 & 0.541 & 0.047 & 0.545 & -0.012 \\
    & Full & 0.676 & 0.676 & 0.495 & 0.495 & 0.674 & 0.071 & 0.434 & 0.428 & 0.415 & 0.295 & 0.382 & 0.380 & 0.413 & 0.138 & 0.220 & 0.201 & 0.553 & 0.171 & 0.559 & 0.167 \\ 
    \midrule 
    \multirow{5}{*}{\shortstack{SAE\\width: 16k\\l0: 116} }
    & \ours & 0.519 & 0.303 & 0.323 & 0.131 & 0.619 & 0.037 & 0.343 & -0.026 & 0.290 & 0.025 & 0.273 & -0.001 & 0.332 & 0.020 & 0.150 & -0.003 & 0.409 & 0.031 & 0.433 & 0.036 \\ & Aura & 0.397 & 0.147 & 0.254 & 0.061 & 0.623 & 0.027 & 0.343 & 0.057 & 0.375 & 0.036 & 0.306 & 0.026 & 0.333 & 0.076 & 0.151 & 0.011 & 0.414 & 0.025 & 0.441 & 0.026 \\ & Range & 0.617 & 0.555 & 0.414 & 0.368 & 0.624 & 0.118 & 0.348 & 0.244 & 0.396 & 0.297 & 0.342 & 0.333 & 0.334 & 0.126 & 0.153 & 0.100 & 0.413 & 0.114 & 0.442 & 0.136 \\ & Adaptive & 0.616 & 0.519 & 0.412 & 0.327 & 0.623 & 0.115 & 0.348 & 0.184 & 0.396 & 0.268 & 0.342 & 0.295 & 0.334 & 0.157 & 0.153 & 0.082 & 0.413 & 0.083 & 0.438 & 0.091 \\ & Full & 0.617 & 0.617 & 0.414 & 0.414 & 0.625 & 0.162 & 0.348 & 0.346 & 0.396 & 0.348 & 0.342 & 0.342 & 0.336 & 0.148 & 0.153 & 0.152 & 0.414 & 0.151 & 0.442 & 0.186 \\ 
    \midrule 
    \multirow{5}{*}{\shortstack{SAE\\width: 65k\\l0: 197}}
    & \ours 
    & 0.572 & 0.222 & 0.435 & 0.113 & 0.639 & 0.024 & 0.420 & -0.038 & 0.273 & 0.026 & 0.275 & -0.010 & 0.391 & 0.028 & 0.230 & -0.005 & 0.556 & 0.025 & 0.667 & 0.013 \\ & Aura & 0.560 & 0.175 & 0.405 & 0.088 & 0.654 & 0.043 & 0.425 & 0.027 & 0.349 & 0.037 & 0.324 & -0.003 & 0.392 & 0.095 & 0.234 & -0.002 & 0.593 & 0.025 & 0.700 & 0.019 \\ & Range & 0.583 & 0.537 & 0.481 & 0.421 & 0.655 & 0.309 & 0.434 & 0.374 & 0.403 & 0.267 & 0.389 & 0.364 & 0.391 & 0.099 & 0.236 & 0.130 & 0.587 & 0.084 & 0.694 & 0.032 \\ & Adaptive & 0.583 & 0.472 & 0.480 & 0.350 & 0.655 & 0.310 & 0.434 & 0.304 & 0.401 & 0.230 & 0.384 & 0.273 & 0.391 & 0.134 & 0.235 & 0.088 & 0.569 & 0.066 & 0.671 & 0.014 \\ & Full & 0.583 & 0.583 & 0.481 & 0.481 & 0.656 & 0.311 & 0.435 & 0.434 & 0.403 & 0.293 & 0.389 & 0.386 & 0.392 & 0.175 & 0.236 & 0.216 & 0.591 & 0.298 & 0.697 & 0.313 \\ 
    \midrule
    \multirow{5}{*}{\shortstack{SAE\\width: 16k\\l0: 285}}
    & \ours & 0.526 & 0.144 & 0.414 & 0.107 & 0.669 & 0.041 & 0.353 & -0.004 & 0.252 & 0.039 & 0.241 & 0.013 & 0.322 & 0.047 & 0.149 & 0.001 & 0.443 & 0.019 & 0.510 & 0.028 \\ & Aura & 0.505 & 0.071 & 0.390 & 0.030 & 0.674 & 0.125 & 0.350 & 0.003 & 0.321 & 0.034 & 0.283 & -0.017 & 0.323 & 0.165 & 0.152 & 0.040 & 0.449 & 0.021 & 0.515 & 0.046 \\ & Range & 0.679 & 0.628 & 0.560 & 0.513 & 0.673 & 0.216 & 0.361 & 0.291 & 0.369 & 0.257 & 0.346 & 0.334 & 0.322 & 0.140 & 0.153 & 0.095 & 0.449 & 0.110 & 0.522 & 0.129 \\ & Adaptive & 0.679 & 0.499 & 0.558 & 0.410 & 0.674 & 0.232 & 0.360 & 0.211 & 0.369 & 0.247 & 0.345 & 0.273 & 0.322 & 0.177 & 0.153 & 0.069 & 0.448 & 0.071 & 0.519 & 0.083 \\ & Full & 0.680 & 0.680 & 0.560 & 0.559 & 0.676 & 0.462 & 0.361 & 0.360 & 0.369 & 0.306 & 0.346 & 0.345 & 0.323 & 0.170 & 0.153 & 0.129 & 0.449 & 0.282 & 0.523 & 0.332 \\
    \bottomrule
  \end{tabular}
  }
\end{table*}

\section{Deepseek Concept Erasure Experiments}
\label{appendix:deepseek_concept_erasure}

On DeepSeek, our method \ours{} is highly competitive on the concept-removal metrics. 
Across all 20 comparisons of $\Delta_{\mathrm{Acc}}$ and $\Delta_{\mathrm{Conf}}$ 
(5 datasets $\times$ 2 metrics $\times$ Base/SAE), 
\ours{} ranks as the best or second-best method in 13 cases.  By contrast, the next strongest method, Range, achieves a top-two ranking in 12 out of 20 comparisons. This demonstrates that \ours{} performs on par with or better than the existing approaches.
Importantly, \ours{} is also the least disruptive method, 
achieving the lowest increase in perplexity (DPPL) in 8 out of 10 DeepSeek experiments 
(all except POS for Base and POS for SAE). 
In one of these two exceptions, \ours{} remains highly competitive with a perplexity increase of 0.269 vs.\ 0.253 for AURA. Overall, \ours{}  combines strong concept-removal performance  with the smallest degradation in language modeling quality on DeepSeek.

Additionally, as shown in Table~\ref{tab:deepseek_concept_erasure}, applying interventions in the \textsc{SAE} representation yields larger $\Delta_{\mathrm{Acc}}$ and $\Delta_{\mathrm{Conf}}$ than in the Base representation across methods: for \ours{}, $\Delta_{\mathrm{Conf}}$ increases on all five datasets and $\Delta_{\mathrm{Acc}}$ increases on 4 out of 5 (slightly lower on \textit{Emotions}), indicating that SAEs promote greater concept separability and thereby enable more effective concept removal.

\section{NER Neurons Activation}
\label{appendix:ner_activations}

As shown in Figure~\ref{fig:range_patterns_ner}, the NER dataset exhibits patterns similar to AG News. Certain neurons in both the base model and SAEs (e.g., leftmost plots) show considerable overlap in their activation distributions across the four classes, indicating limited class discrimination. In contrast, neurons in the middle and right columns reveal more separable activation patterns.

\begin{table*}[t!]
  \centering
  \caption{Concept Erasure Detailed metrics \(D_{\mathrm{Acc}}\), \(D'_{\mathrm{Acc}}\), \(D_{\mathrm{Conf}}\), and \(D'_{\mathrm{Conf}}\) for DeepSeek-R1.}
  \label{tab:deepseek_concept_erasure_details}
  \resizebox{\textwidth}{!}{%
  \begin{tabular}{l l
                  |cccc
                  |cccc
                  |cccc
                  |cccc
                  |cccc
                  }
    \toprule
    \textbf{Type} & \textbf{Method}
      & \multicolumn{4}{c}{\textbf{POS}}
      & \multicolumn{4}{c}{\textbf{AG News}}
      & \multicolumn{4}{c}{\textbf{Emotions}}
      & \multicolumn{4}{c}{\textbf{DBpedia}}
      & \multicolumn{4}{c}{\textbf{NER}} \\
    \cmidrule(lr){3-6} \cmidrule(lr){7-10}
    \cmidrule(lr){11-14} \cmidrule(lr){15-18} \cmidrule(lr){19-22}
    & 
      & \small$D_{\mathrm{Acc}}\uparrow$
      & \small$D'_{\mathrm{Acc}}\downarrow$
      & \small$D_{\mathrm{Conf}}\uparrow$
      & \small$D'_{\mathrm{Conf}}\downarrow$
      & \small$D_{\mathrm{Acc}}\uparrow$
      & \small$D'_{\mathrm{Acc}}\downarrow$
      & \small$D_{\mathrm{Conf}}\uparrow$
      & \small$D'_{\mathrm{Conf}}\downarrow$
      & \small$D_{\mathrm{Acc}}\uparrow$
      & \small$D'_{\mathrm{Acc}}\downarrow$
      & \small$D_{\mathrm{Conf}}\uparrow$
      & \small$D'_{\mathrm{Conf}}\downarrow$
      & \small$D_{\mathrm{Acc}}\uparrow$
      & \small$D'_{\mathrm{Acc}}\downarrow$
      & \small$D_{\mathrm{Conf}}\uparrow$
      & \small$D'_{\mathrm{Conf}}\downarrow$
      & \small$D_{\mathrm{Acc}}\uparrow$
      & \small$D'_{\mathrm{Acc}}\downarrow$
      & \small$D_{\mathrm{Conf}}\uparrow$
      & \small$D'_{\mathrm{Conf}}\downarrow$ \\
    \midrule 
    \multirow{5}{*}{Base} 
    & \ours 
      & 0.198 & 0.013 & 0.199 & -0.011
      & 0.193 & 0.009 & 0.363 & -0.021
      & 0.106 & 0.007 & 0.162 & -0.029
      & 0.160 & 0.005 & 0.242 & -0.008
      & 0.095 & 0.007 & 0.133 & -0.010 \\
    
    & Aura  
      & 0.081 & 0.032 & 0.103 & 0.034
      & 0.029 & 0.014 & 0.125 & 0.054
      & 0.099 & 0.032 & 0.179 & 0.033
      & 0.241 & 0.030 & 0.348 & 0.067
      & 0.051 & 0.013 & 0.074 & 0.003 \\
    
    & Range 
      & 0.067 & 0.011 & 0.288 & 0.065
      & 0.303 & 0.130 & 0.687 & 0.469
      & 0.148 & 0.070 & 0.315 & 0.164
      & 0.181 & 0.032 & 0.347 & 0.021
      & 0.144 & 0.054 & 0.379 & 0.123 \\
    
    & Adaptive 
      & 0.060 & 0.010 & 0.140 & 0.001
      & 0.191 & 0.069 & 0.519 & 0.243
      & 0.121 & 0.050 & 0.237 & 0.077
      & 0.136 & 0.021 & 0.228 & -0.016
      & 0.121 & 0.031 & 0.236 & 0.039 \\
    
    & Full 
      & 0.062 & 0.008 & 0.291 & 0.109
      & 0.287 & 0.171 & 0.691 & 0.575
      & 0.147 & 0.078 & 0.318 & 0.183
      & 0.183 & 0.047 & 0.353 & 0.064
      & 0.127 & 0.060 & 0.371 & 0.199 \\ 
    
    \midrule 
    \multirow{5}{*}{SAE}
    & \ours 
      & 0.325 & 0.018 & 0.338 & 0.011
      & 0.507 & 0.020 & 0.674 & -0.134
      & 0.098 & 0.006 & 0.339 & -0.134
      & 0.660 & 0.008 & 0.573 & -0.085
      & 0.581 & 0.014 & 0.838 & -0.039 \\
    
    & Aura 
      & 0.424 & 0.010 & 0.455 & 0.043
      & 0.522 & 0.008 & 0.688 & -0.086
      & 0.098 & 0.009 & 0.339 & -0.071
      & 0.664 & 0.012 & 0.576 & -0.198
      & 0.604 & 0.015 & 0.869 & -0.014 \\
    
    & Range 
      & 0.420 & 0.067 & 0.479 & 0.057
      & 0.497 & 0.044 & 0.644 & -0.202
      & 0.098 & 0.008 & 0.329 & -0.329
      & 0.515 & 0.017 & 0.347 & -0.299
      & 0.596 & 0.055 & 0.855 & 0.025 \\
    
    & Adaptive 
      & 0.375 & 0.036 & 0.423 & -0.008
      & 0.465 & 0.032 & 0.607 & -0.224
      & 0.098 & 0.005 & 0.328 & -0.324
      & 0.499 & 0.011 & 0.337 & -0.294
      & 0.546 & 0.045 & 0.793 & 0.004 \\
    
    & Full 
      & 0.422 & 0.177 & 0.481 & 0.174
      & 0.501 & 0.301 & 0.652 & 0.174
      & 0.098 & 0.008 & 0.329 & -0.415
      & 0.520 & 0.022 & 0.350 & -0.326
      & 0.603 & 0.303 & 0.867 & 0.378 \\
    
    \bottomrule
  \end{tabular}
  }
\end{table*}

\FloatBarrier

\section{Details of Experiments}
\label{app:details_metrics}

The detailed metrics (\(D_{\mathrm{Acc}}\), \(D'_{\mathrm{Acc}}\), \(D_{\mathrm{Conf}}\), and \(D'_{\mathrm{Conf}}\)) before subtraction for Gemma-2-2b are reported in Table~\ref{tab:gemma_concept_erasure_details}, and those for DeepSeek-R1 are presented in Table~\ref{tab:deepseek_concept_erasure_details}.

\section{Implementation Details}
\label{appendix:implementation_details}

\stitle{Hyperparameters.}
Range Masking, Adaptive Dampening, and Full Masking each rely on a saliency threshold hyperparameter that determines the top-\(p\%\) of neurons considered most relevant to a given concept. For Gemma, we set \(p = 0.3\) for POS, AG News and NER, \(p = 0.2\) for Emotions and DBpedia. For DeepSeek, we use \(p = 0.3\), \(0.4\), \(0.1\), \(0.2\), and \(0.4\) for POS, AG News, Emotions, DBpedia, and NER, respectively. In contrast, AURA~\cite{aura2024} and our method, {\ours}, do not require this hyperparameter. Both apply interventions across all neurons in the selected layer, avoiding the need to tune or justify a saliency cutoff.
However, because SAE neurons typically activate on only a small subset of inputs for any given concept, we introduce an activation-frequency threshold \(\tau\) to ensure reliability. Specifically, for each SAE neuron \(h^l_j\), and each concept \(c_i\), let \(\mathcal{X}_{c_i}\) be the set of corresponding input samples. We define the firing frequency of \(h^l_j\) on \(c_i\) as
\[
f_{j,i}
=
\frac{\bigl|\{\,x\in \mathcal{X}_{c_i}  : h^l_j(x) > 0\}\bigr|}{|\mathcal{X}_{c_i}|}.
\]
We exclude neuron \(h^l_j\) from all concept-erasure methods if \(f_{j,i} < \tau\), as sparse activations preclude meaningful intervention. In all experiments, we set \(\tau = 0.1\).

\stitle{Histogram-based KDE.}
To estimate the densities \(f_{h^l_j \mid c_i}(x)\) required by \ours, we use a kernel density estimation (KDE) implemented via a histogram-based approximation that preserves accuracy while greatly improving efficiency. During training, each neuron–concept activation distribution is discretized into \(B\) uniform-width bins (we use \(B=2048\)) and we store the bin centers, counts, bandwidths, and normalization constants. At inference, a query activation \(x\) is evaluated only against these \(B\) centers rather than all \(N\) training activations. Let \(F\) be the number of neurons and \(Q\) the number of query points; this reduces complexity from \(\mathcal{O}(F N Q)\) time and \(\mathcal{O}(F N)\) memory for naïve KDE to \(\mathcal{O}(F B Q)\) and \(\mathcal{O}(F B)\), respectively, yielding an approximate \(N/B\)-fold improvement in both inference speed and memory usage (\(B \ll N\)).

\stitle{Intervention Location and Scope.}
All interventions are applied at a consistent computation point (immediately before the language-model head) to ensure comparability across models and methods. Specifically, interventions are introduced after the MLP and residual addition in the final transformer block. For SAEs, they are applied directly after the SAE activation nonlinearity. To isolate causal effects on prediction, interventions are restricted to the token corresponding to the model’s output label. This design allows precise measurement of each concept-erasure method’s influence on the final decision while avoiding confounding effects from earlier tokens.

\stitle{Dataset Statistics.}  
We report the dataset sizes and splits used in all experiments to ensure transparency and reproducibility. 
The AG News dataset contains 120{,}000 training and 7{,}600 test examples. The DBpedia dataset includes 100{,}000 training and 25{,}000 test examples, both randomly sampled using a fixed seed to ensure reproducibility. 
For Emotions, we use 16{,}000 training and 2{,}000 test examples. 
The POS Tagging and NER datasets each consist of 100{,}000 training and 10{,}000 test examples, also sampled with a fixed random seed.

\stitle{Model Specifications.}
As described in the main text, we use the Gemma-2 model with 2\,billion parameters (Hugging Face name: \texttt{google/gemma-2-2b}). 
For DeepSeek, we employ the DeepSeek-R1 model, specifically the distilled variant based on Llama-8B with 8\,billion parameters (Hugging Face name: \texttt{deepseek-ai/ DeepSeek-R1-Distill-Llama-8B}). 

\stitle{Computation Details.}
All experiments were conducted using the university's high-performance computing (HPC) cluster managed via the Slurm workload manager. We used NVIDIA A100 (40\,GB) and V100 (32\,GB) GPUs for activation collection and intervention process.
Each job was allocated 1 GPU, 1–10 CPU cores, and 50\,GB of RAM.
% The total computational budget for all experiments was approximately 250 GPU-hours, including training, evaluation, and ablation studies.

\stitle{Result Reliability.}
All quantitative results reported in the concept-erasure tables correspond to a single representative run. To verify the stability of our findings, we reran selected datasets (e.g., AG News) multiple times and observed identical outcomes and consistent relative rankings across methods. For instance, the top-performing method in the initial run remained the best in repeated trials. These consistent results indicate that the reported values are reliable and reproducible.

\stitle{Package Implementation and Parameter Settings.}
Our implementation leverages several standard Python and deep learning libraries. The Hugging Face Transformers package was used to load and run language models (AutoTokenizer, AutoModelForCausalLM). The transformer\_lens and sae\_lens libraries were employed for model inspection and sparse autoencoder integration.
The scikit-learn library was used for evaluation metrics such as ROC-AUC, while pandas and numpy supported data processing. Default configurations were used for all libraries, with reproducibility ensured by fixing random.seed(42).

\begin{figure}[t!]
  \centering
  \includegraphics[width=0.9\linewidth]{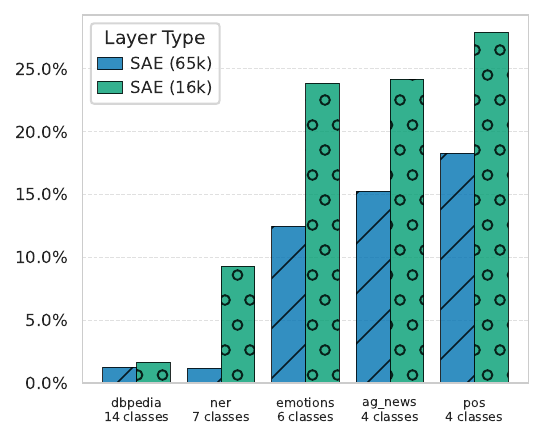}
  \caption{65k SAEs show lower overlap than 16k, further supporting that greater capacity enables more distinct neuron-to-concept mappings.}
  \vspace{-12pt}
  \label{fig:overlap_SAE_percentage}
\end{figure}

\begin{figure*}[t]
    \centering
        \includegraphics[width=0.48\textwidth]{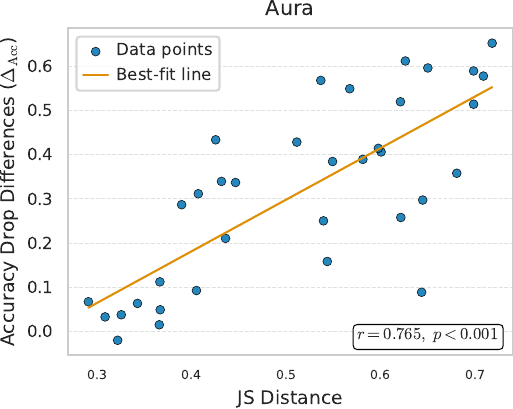}
    \includegraphics[width=0.48\textwidth]{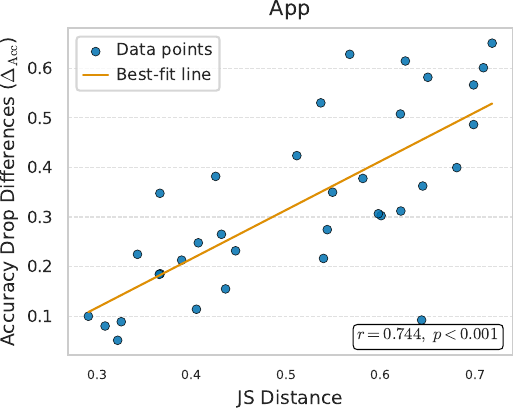}
    \includegraphics[width=0.48\textwidth]{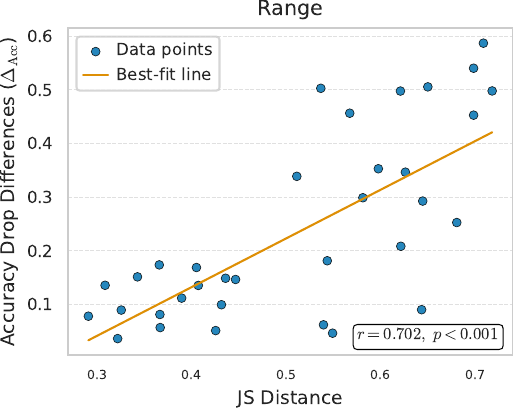}
    \includegraphics[width=0.48\textwidth]{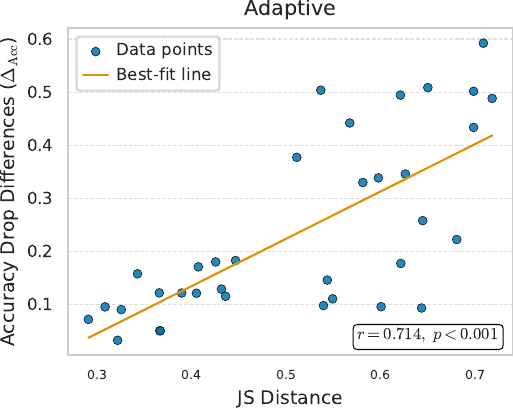}
    \caption{JS distance vs. erasure ability for all partial erasure methods.}
    \label{fig:js_vs_erasure_methods}
    \vspace{-15pt}
\end{figure*}

\section{All SAE Neurons Analysis}
\label{app:sae_analysis}

While our salient neuron analysis provides valuable insight into the most strongly responding neurons, it considers only a narrow slice of the activation space (specifically, the top 80 neurons per concept). This limited scope may miss neurons that, although not highly ranked by mean activation, are still consistently active across multiple concepts and contribute to polysemanticity. To address this limitation, we broaden our analysis to include all neurons that exhibit non-zero activation for any concept, offering a more comprehensive view of concept overlap beyond the most salient neurons. This extended analysis is conducted only for SAEs, since in the dense base model, all neurons are active across all inputs, rendering such overlap statistics uninformative. The results are shown in Figure~\ref{fig:overlap_SAE_percentage}, which reports the intersection-over-union of active neurons across concepts. By capturing both highly active and more subtly engaged neurons, this analysis reveals a complete picture of polysemantic behavior. Consistent with our earlier findings shown in Figure~\ref{fig:overlap_percentage}, we observe that higher-capacity SAEs (e.g., 65k dimensions) exhibit lower neuron overlap than their lower-capacity counterparts. Together, Figures~\ref{fig:overlap_percentage} and~\ref{fig:overlap_SAE_percentage} demonstrate that while SAEs significantly reduce polysemanticity, they do not eliminate it entirely. Polysemantic neurons remain present, albeit to a lesser extent.

%While our salient neuron analysis provides valuable insights into the most strongly responding neurons, it captures only a subset of all neuorns (in our case only the top 50 neurons). Many neurons may be consistently active across multiple concepts without ranking among the most highly activated ones. To address this limitation, Figure~\ref{fig:overlap_SAE_percentage} expands our analysis for SAEs to encompass all neurons that become active for any given concept. In this approach, we collect all neurons with non-zero activation values for each concept, regardless of activation magnitude. We then compute the overlap by determining how many active neurons are shared across all concepts relative to the total number of unique active neurons across the dataset.

\section{Concept Erasure and JS Distance}
\label{app:correlations}

As illustrated in Figure~\ref{fig:js_vs_erasure_methods}, all partial concept erasure methods exhibit a strong Pearson correlation with our JS separability score metric. Specifically, each method achieves a correlation coefficient greater than 0.7 with statistically significant \( p \)-values (\( p < 0.01 \)), indicating a robust relationship between concept separability and erasure precision.
These results suggest that partial erasure techniques can effectively leverage the separability inherent in activation representations to suppress the targeted concept while minimizing interference with unrelated ones. Among the four partial erasure approaches evaluated, AURA and \ours{} demonstrate the highest correlations (0.765 and 0.744, respectively), highlighting their superior ability to exploit distributional distinctions between concept activations. We attribute this performance advantage to the fact that both AURA and \ours{} explicitly model the distributions of both target and auxiliary concepts as discussed in Subsection~\ref{title:aura_and_ours_analysis} (Comparative Analysis of APP and AURA).
%this joint modeling enables these methods to precisely localize and dampen concept-specific activation regions, thereby achieving higher erasure fidelity with minimal side effects.

\section{JS Separability: DeepSeek vs. Gemma}
\label{app:all_js}

Table \ref{tab:js_gemma_deepseek} presents the JS separability scores computed for both the Gemma-2-2B and DeepSeek-R1 models across all datasets. As indicated in the table, every SAE variant consistently achieves higher separability scores than its respective base model. This consistent improvement confirms that the incorporation of SAEs enhances the distinction between concept activation distributions, leading to more interpretable internal representations. Moreover, when comparing SAEs with the same capacity, those with higher sparsity (corresponding to lower $l_0$ values) exhibit greater separability. These results confirms that sparsity plays a critical role in improving the distinctness of concept distributions.

\begin{table*}[t!]
\centering
\small
\setlength{\tabcolsep}{5pt}
\renewcommand{\arraystretch}{1.2}
\resizebox{\textwidth}{!}{%
\begin{tabular}{p{0.8cm} p{1.6cm} p{1.8cm} p{6.8cm} ccc}
\toprule
\textbf{ID} & \textbf{Primary Class} & \textbf{Related Class} & \textbf{Sentence (excerpt)} & \textbf{Base 1621} & \textbf{SAE-16k 332} & \textbf{SAE-65k 5997} \\
\midrule
1 & Business & Science/Tech &Microsoft, Dell big winners of Air Force deal ... for software services worth \$500M & $-12.18$ & $20.15$ & $61.67$ \\
2 & Business & Sports & Retailers bet on poker's rising popularity ... playing cards and chips ... holiday shopping season & $0.52$ & $26.31$ & $45.32$ \\
3 & Business & World & Intellectual property rights ... US companies in China ... WTO commitments & $-4.27$ & $12.16$ & $79.13$ \\
4 & Sports & World & Doping cases hit record ... 24 athletes ousted for drug-related violations ... Olympics ... no US team member flunked a test & $0.64$ & $20.24$ & $49.52 $ \\
5 & Sports & Science/Tech & Suggs to practice ... Cleveland Browns running back Lee Suggs cleared by doctors to return ... still some haziness about his medical condition & $5.14$ & $26.99$ & $54.39$ \\
6 & World & Science/Tech & Ivan skirts Grand Cayman, flooding homes ... winds near 250 km/h ... uprooting trees, bursting canals, and flooding homes as it churned toward Cuba & $-12.35$ & $19.02$ & $62.22$ \\
\bottomrule
\end{tabular}
}
\caption{Example AG News sentences illustrating cross-class activation patterns. Each sentence’s primary class is shown alongside a semantically related secondary class. Activation values reflect the response of three separable neurons identified in Figure~\ref{fig:range_patterns_agnews}.}
\label{tab:cross_class_agnews}
\end{table*}

\begin{table}[t!]
\centering
\resizebox{\columnwidth}{!}{%
\begin{tabular}{l|c|c|c|c|c}
\toprule
& \textbf{POS} & \textbf{AG News} & \textbf{Emotions} & \textbf{DBpedia} & \textbf{NER} \\
\midrule
\multicolumn{6}{c}{\textbf{Gemma}} \\
\midrule
\textbf{Base}
 & 0.343 & 0.366 & 0.322 & 0.405 & 0.308 \\
\makecell[l]{\textbf{SAE}\\(width\_16k/l0\_116)}
 & 0.539 & 0.650 & 0.431 & 0.621 & 0.581 \\
\makecell[l]{\textbf{SAE}\\(width\_16k/l0\_285)}
 & 0.425 & 0.567 & 0.389 & 0.543 & 0.511 \\
\makecell[l]{\textbf{SAE}\\(width\_65k/l0\_93)}
 & \textbf{0.600} & \textbf{0.709} & \textbf{0.446} & \textbf{0.680} & \textbf{0.621} \\
\makecell[l]{\textbf{SAE}\\(width\_65k/l0\_197)}
 & 0.549 & 0.626 & 0.407 & 0.644 & 0.537 \\
\midrule
\multicolumn{6}{c}{\textbf{DeepSeek}} \\
\midrule
\textbf{Base}
 & 0.367 & 0.366 & 0.291 & 0.436 & 0.325 \\
\textbf{SAE}
 & \textbf{0.597} & \textbf{0.698} & \textbf{0.643} & \textbf{0.718} & \textbf{0.698} \\
\bottomrule
\end{tabular}%
}
\caption{JS separability score comparison across datasets for Gemma and DeepSeek. Bold values indicate the best score per dataset.}
\label{tab:js_gemma_deepseek}
\vspace{-10pt}
\end{table}

\FloatBarrier

\section{Concept Distribution Analysis}
\label{app:distribution_analysis}
We analyze concept-conditional activation distributions on the AG News dataset, focusing on three neurons with high class separability in Figure~\ref{fig:range_patterns_agnews}: Base Neuron 1621, SAE-16k Neuron 332, and SAE-65k Neuron 5997.

To examine how these neurons encode secondary topical cues, consider the following example from the AG News dataset:
\begin{quote}
\textit{``Microsoft, Dell big winners of Air Force deal ... for software services ... worth \$500 million ...''}
\end{quote}
The primary label of this sentence is Business, though it clearly involves elements of Science/Tech. The per-neuron activations for this sentence are:

\begin{itemize}
    \item Base Neuron 1621: $-12.1797$
    \item SAE-16k Neuron 332: $20.1485$
    \item SAE-65k Neuron 5997: $61.6663$
\end{itemize}

Across all three neurons, the activations fall within the Business class density in Fig.~\ref{fig:range_patterns_agnews} but are shifted toward the Science/Tech density. This pattern is consistent with polysemantic encoding; the neurons primarily represent the Business topic (capturing the correct class) while remaining sensitive to  Science/Tech signals (e.g., “Microsoft,” “software”), which pulls the activation toward the Science/Tech mode without leaving the Business distribution.

Table~\ref{tab:cross_class_agnews} extends the analysis to six mixed-topic sentences. In every case, the three neurons produce activations that sit within the primary class distribution yet are shifted toward the related class. %, mirroring the overlap in Fig.~\ref{fig:range_patterns_agnews}.
%Concretely, Science/Tech cues such as “Microsoft,” “software,” and numerical magnitudes (rows 1 and 6) elicit sizeable responses in SAE-65k-5997 and SAE-16k-332; legal/geo-political terms (“WTO,” “China,” row 3) likewise nudge activations toward the World distribution; and medical/anti-doping vocabulary (rows 4–5) shifts Sports examples toward Sci/Tech.
Absolute scales differ across neurons, so values should be interpreted relatively against each neuron’s class-conditional densities in Fig.~\ref{fig:range_patterns_agnews} rather than compared across neurons. Overall, the pattern supports a polysemantic account; these neurons track the primary topic while remaining sensitive to secondary topical signals.

\end{document}